\documentclass[letterpaper, 10 pt, conference]{ieeeconf}  

\IEEEoverridecommandlockouts                              

\usepackage{algorithm}
\usepackage{algorithmic}

\usepackage{times}
\usepackage{epsfig}
\usepackage{graphicx}
\usepackage{amsmath}
\usepackage{amssymb}

\usepackage{hyperref}
\usepackage{booktabs}
\usepackage{multirow}
\usepackage{makecell}
\usepackage{colortbl}
\usepackage{tabulary}
\usepackage[table]{xcolor}
\usepackage{afterpage}
\usepackage{lipsum} 
\usepackage[absolute,overlay]{textpos}
\usepackage{adjustbox}
\usepackage{cuted}  
\usepackage{hyperref}
\usepackage{amsmath}
\usepackage{marvosym}

\usepackage{color} 

\usepackage{subcaption}
\usepackage{pifont}
\definecolor{darkbrown}{RGB}{135,  60,   0}
\definecolor{orrange}{RGB}{255, 120,  50}
\definecolor{pink}{RGB}{255, 192, 203}
\definecolor{yellow}{RGB}{255, 255,   0}
\definecolor{blue}{RGB}{  0, 150, 245}
\definecolor{cyan}{RGB}{ 0, 255, 255}
\definecolor{green}{RGB}{0, 175,   0}

\usepackage{float}
\usepackage{listings}

\title{\LARGE \bf
SliceOcc:  Indoor 3D Semantic Occupancy Prediction with Vertical Slice Representation
}

\author{Jianing Li$^{1}$, Ming Lu$^{2}$, Hao Wang$^{2}$, Chenyang Gu$^{2}$, Wenzhao Zheng$^{3}$, Li Du$^{1}$, Shanghang Zhang$^{2}$
\thanks{$^{1}$Jianing Li and Li Du are with the School of Electronic Sci
ence and Engineering, Nanjing University, Nanjing, China.}  \thanks{$^{2}$Ming Lu, Hao Wang, Chenyang Gu, Shanghang Zhang are with the School of Computer Science, Peking University, Beijing, China.}\thanks{$^{3}$Wenzhao Zheng is with the Department of Electrical Engineering and Computer Sciences,  University of California, Berkeley, Berkeley, USA.}
\thanks{$^{\text{\Letter}}$Corresponding Author: ldu@nju.edu.cn; shanghang@pku.edu.cn}
}

\begin{document}

\maketitle


\begin{abstract}
3D semantic occupancy prediction is a crucial task in visual perception, as it requires the simultaneous comprehension of both scene geometry and semantics. It plays a crucial role in understanding 3D scenes and has great potential for various applications, such as robotic vision perception and autonomous driving. Many existing works utilize planar-based representations such as Bird’s Eye View (BEV) and Tri-Perspective View (TPV). These representations aim to simplify the complexity of 3D scenes while preserving essential object information, thereby facilitating efficient scene representation. However, in dense indoor environments with prevalent occlusions, directly applying these planar-based methods often leads to difficulties in capturing global semantic occupancy, ultimately degrading model performance. In this paper, we present a new vertical slice representation that divides the scene along the vertical axis and projects spatial point features onto the nearest pair of parallel planes. To utilize these slice features, we propose SliceOcc, an RGB camera-based model specifically tailored for indoor 3D semantic occupancy prediction. SliceOcc utilizes pairs of slice queries and cross-attention mechanisms to extract planar features from input images. These local planar features are then fused to form a global scene representation, which is employed for indoor occupancy prediction. Experimental results on the EmbodiedScan dataset demonstrate that SliceOcc achieves a mIoU of 15.45\% across 81 indoor categories, setting a new state-of-the-art performance among RGB camera-based models for indoor 3D semantic occupancy prediction. Code is available at: \textcolor{magenta}{https://github.com/NorthSummer/SliceOcc}.

\end{abstract}

\section{Introduction}
3D scene understanding is a critical task across various artificial intelligence domains, including robotic vision perception and autonomous driving. To comprehensively capture both the geometric and semantic aspects of 3D scenes, the concept of 3D semantic occupancy was first introduced in the context of autonomous driving \cite{huang2023tri, wang2023openoccupancy, wei2023surroundocc, zhang2023occformer, zuo2023pointocc, tang2024sparseocc, 10610625, zheng2025occworld, huang2025gaussianformer}. This task involves dividing the visual space into fixed cubic grids. It predicts the occupancy status and semantic category for each grid cell, providing a comprehensive understanding of driving scenarios in both geometry and semantics. Existing methods for 3D occupancy prediction have mainly explored two types of representations: voxel-based and planar-based. Voxel-based methods typically involve a 2D-to-3D lifting process \cite{zhang2023occformer, tang2024sparseocc} or point voxelization \cite{wang2023openoccupancy}, mapping input data (such as images or point clouds) onto a voxel space, followed by the use of 3D convolutions to extract voxel features. In contrast, planar-based methods \cite{huang2023tri, zuo2023pointocc, 10610625} project input data onto planes, such as Bird's Eye View (BEV) or Tri-Perspective View (TPV) planes, generating planar features, which are subsequently used to derive voxel features for 3D semantic occupancy prediction.

\begin{figure}[t]
    \centering
    \includegraphics[width=0.85\linewidth, height=6.3cm]{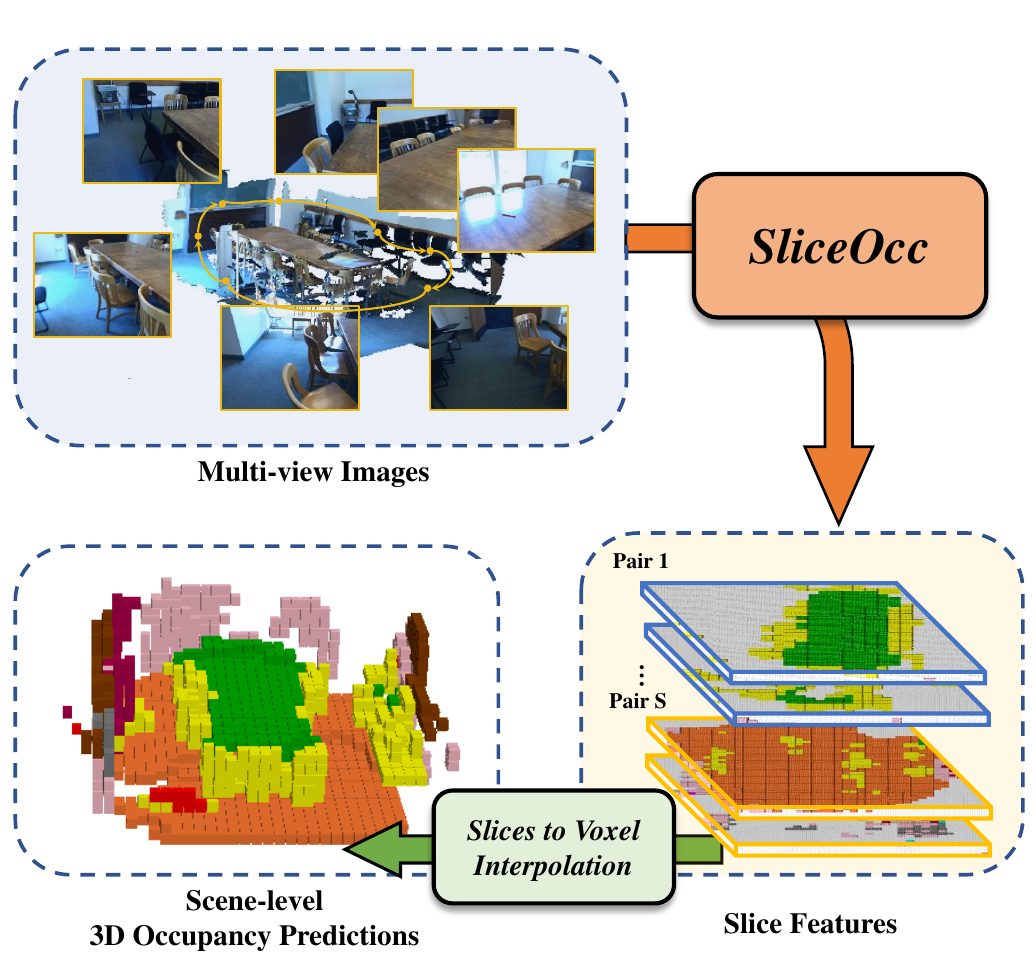}
    \caption{Given a set of multi-view images from an indoor scene, SliceOcc vertically slices the scene to generate pairs of slice features, which are then interpolated to form voxel features for 3D semantic occupancy prediction.}
    \label{fig:cover}
\end{figure}

While these approaches have shown promising results in outdoor autonomous driving scenarios, their application to indoor environments remains underexplored. The significant disparities between indoor and outdoor environments pose distinct challenges when directly applying outdoor methodologies to indoor settings. From a data perspective, autonomous driving scenarios generally have a vision-centric distribution, with the prediction space confined to a fixed area around the ego vehicle. In contrast, indoor scenes in robotic vision systems often involve extreme variations in orientation and pose, covering the entire environment rather than a localized area. Regarding scene composition, outdoor environments often contain many moving objects spread out over significant distances. Conversely, indoor scenes are characterized by static objects densely packed within confined spaces, resulting in a much higher likelihood of occlusions compared to outdoor environments. Given these contrasting conditions, directly applying outdoor methods to indoor scenes may result in degraded performance due to the dense occlusions inherent in indoor environments.


In this paper, we present SliceOcc, an advanced planar-based method for 3D semantic occupancy prediction that utilizes image inputs and implements a slicing strategy to effectively interpret indoor scenes. Since indoor objects such as tables, chairs, and beds mostly have horizontal surfaces and consistent vertical orientation, we divide the indoor scene uniformly along the vertical axis. Each sliced part of the scene is represented using a pair of planes. We utilize SliceOcc to combine the geometric and semantic information of the sliced scene onto these planes, which helps us derive the features of the sliced scene, known as slice features. We then interpolate the slice features to recover vertical information, ultimately creating voxel features for the entire space. Fig. \ref{fig:cover} illustrates the abovementioned process.

Specifically, SliceOcc is a transformer-based model that uses pairs of slice queries to obtain the slice features of the scene. These queries are initialized using 3D anchors and processed through a series of cross-attention mechanisms. For each slice query, sliced spatial cross-attention is employed to project image features onto the slice plane and its surrounding space. Planar cross-attention enables neighboring slice queries within the same slice level to interact, improving consistency in the vertical direction. The voxel features for the entire scene are formed by concatenating the slice features generated by SliceOcc and then interpolating them pair-wise along the vertical direction. These voxel features are finally passed through an occupancy head for 3D semantic occupancy prediction.

Our main contributions are as follows:

\begin{itemize}
\item We introduce SliceOcc, a 3D semantic occupancy prediction model designed for indoor scenes, which takes multi-view images as input.
\item We propose a vertical slice scene representation method tailored to indoor environments. By leveraging attention mechanisms, we facilitate effective information interaction between planes and images, as well as between neighboring planes, thereby enhancing model performance in complex indoor settings.
\item We conduct comprehensive experiments on the EmbodiedScan dataset, demonstrating that our method achieves a mIoU of 15.45\% across 81 object categories. This performance reaches state-of-the-art levels among camera-based models and is comparable to the results obtained by leading depth-based models.
\end{itemize}
\vspace{-3pt}
\section{Related Work}
\subsection{Planar-based scene representation}
Common methods for representing scenes include planar-based and voxel-based approaches, each capturing 3D spatial information differently. Unlike voxel-based methods, which represent the entire scene in dense 3D grids, planar-based approaches typically focus on extracting information from key planes (such as BEV and TPV) within the scene. These approaches use planar features to capture the characteristics of the entire environment implicitly. Recently, many vision-based autonomous driving approaches have employed the BEV plane to effectively represent vision-centric driving scenes, achieving significant success in tasks like 3D object detection and map segmentation. In terms of obtaining BEV features, some methods directly project image features into frustum space using depth information \cite{philion2020lift, huang2021bevdet, li2023bevdepth, park2022time, chi2023bev}. In contrast, others utilize transformer architectures to query relevant information from image features implicitly \cite{li2022bevformer, liu2022petr, wang2023exploring}. On the other hand, TPV represents the scene using three orthogonal planar features, providing more comprehensive geometric information, making it more widely applied in denser tasks such as object-level or human facial surface reconstruction \cite{wang2023pet, wang2023alto, ma2023otavatar, zou2024triplane}.

\subsection{3D semantic occupancy prediction}
3D semantic occupancy prediction was initially designed to assist in achieving a denser 3D perception paradigm for autonomous driving scenarios. Leveraging the concept of occupancy \cite{mescheder2019occupancy, peng2020convolutional}, existing methods divide the entire scene into regular 3D grids and predict semantic information for each grid, aiming to obtain a consistent geometric and semantic estimate of the entire scene. TPVFormer \cite{huang2023tri} is the first to initiate academic discussions on how to estimate 3D semantic occupancy. It utilizes TPV representations, where image-TPV cross attention is employed to acquire features projected from spatial points to three orthogonal planes.  SurroundOcc \cite{wei2023surroundocc} defines queries at the 3D volume space and uses image-volume cross attention to directly obtain voxel features. OpenOccupancy \cite{wang2023openoccupancy} utilizes the LSS method to lift image features into a low-dimensional 3D volume space. There are also methods \cite{pan2024renderocc, huang2024selfocc, zhao2024hybridocc} that model 3D grids from semantic and density perspectives and render novel views through neural fields to achieve a self-supervised manner of occupancy prediction. For indoor scenes, EmbodiedScan \cite{wang2024embodiedscan} has released a large-scale semantic occupancy prediction benchmark. ISO \cite{yu2024monocular} uses monocular images and predicted depths to estimate the 3D semantic occupancy. EmbodiedOcc \cite{wu2024embodiedocc} employs 3D Gaussians for the representation of indoor scene and dynamically updates these Gaussians states through continuous scanning processes.

\begin{figure*}[t]
    \centering
    \includegraphics[width=0.92\textwidth, height=8.5cm]{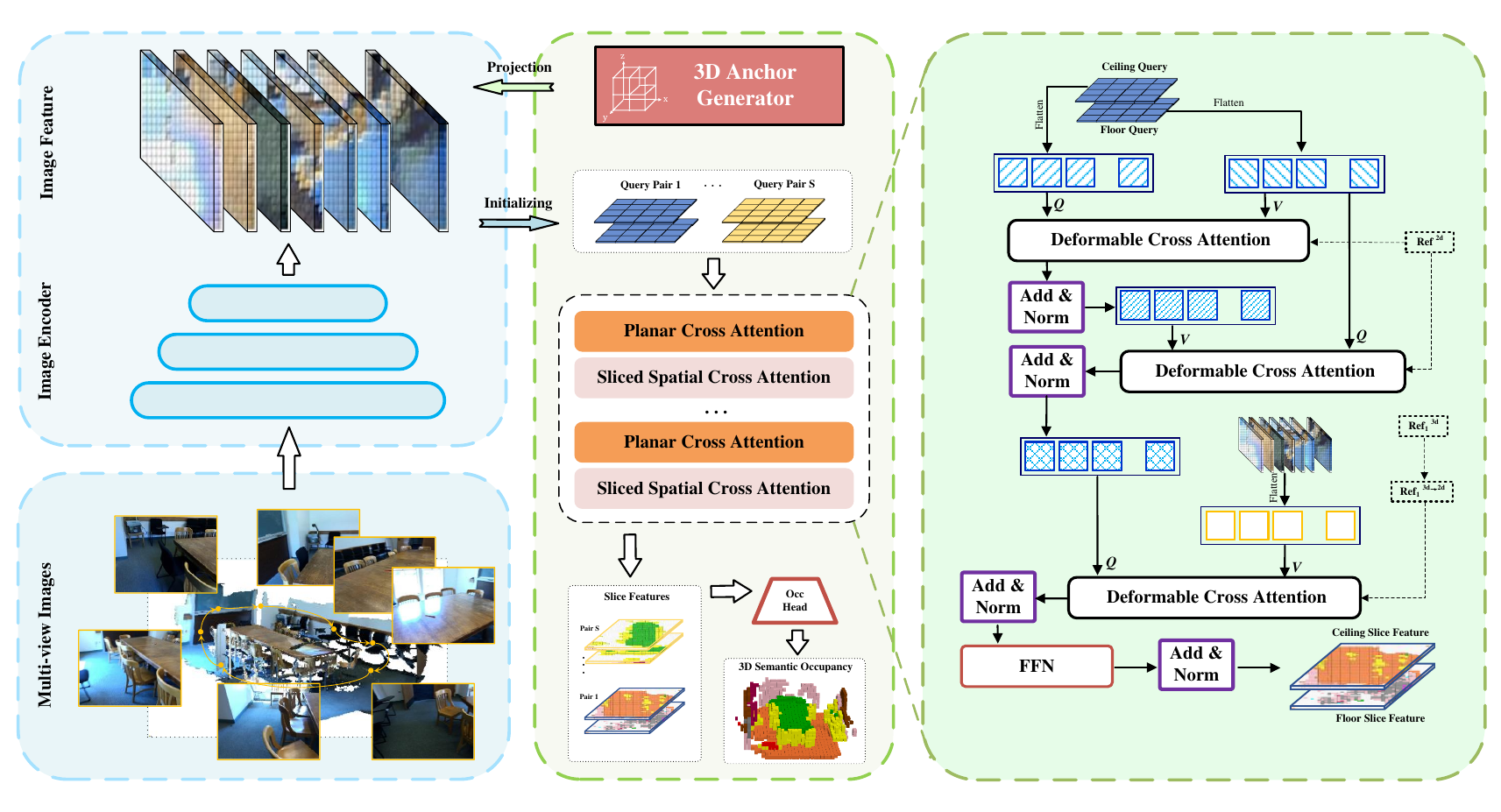}
    \caption{The framework of SliceOcc. SliceOcc consists of an image encoder and an occupancy decoder. We use an image backbone as the encoder to extract image features. In the occupancy decoder, we employ SSCA (sliced spatial cross attention) to enable interaction between the slice features and the image features, allowing for information exchange. We then design PCA (planar cross attention) to refine the slice features. Finally, we concatenate the slice features along the height dimension and make predictions for the entire scene's semantic occupancy.}
    \label{fig:framework}
\end{figure*}
\vspace{-3pt}
\section{Method}
\subsection{Problem definition}

In an indoor scene, we have N sampled images along with their corresponding camera intrinsics ${ K_s \in \mathbb{R}^{3\times3} }$ and camera extrinsics ${ E_s=(\mathbf{R_s}|\mathbf{T_s}) \in \mathbb{R}^{4\times3} }$, where $s$ ranges from 1 to N, representing the image index. We establish a global coordinate system within the scene to derive the 3D semantic occupancy. The scene is divided into uniform grids of equal width, length, and height so that every spatial point $p = (x, y, z)$ corresponds to an occupancy grid $\mathbf{O}_{x,y,z}$. The model takes as input $N$ multi-view images along with the corresponding camera parameters and predicts both the occupancy status (empty or occupied) and the dominant semantic category for each grid under the supervision of semantic occupancy labels.

In our approach, we aim to forecast 3D semantic occupancy using a vertical slice representation. We divide the scene, with a total height of $H$, into $S$ parts along the vertical direction. Each sliced scene is bounded by a floor plane and a ceiling plane. We can take any spatial point $p = (x, y, z)$ and project it onto the two adjacent planes (floor and ceiling). This shows us that we can reconstruct the features of any spatial point from the features projected onto these adjacent planes. Thus, we propose SliceOcc, which is designed to query slice features using an attention mechanism.
\subsection{Framework of SliceOcc}
The overall architecture of SliceOcc is illustrated in Fig. \ref{fig:framework}. We employ an image backbone as the encoder to extract image features $\mathcal{F}$ from the multi-view images. Subsequently, we grid-sample using 3D anchors from the image features and add learnable height embeddings to initialize the slice queries for the occupancy decoder. In the occupancy decoder, we introduce two attention blocks: the Planar Cross-Attention block (PCA) and the Sliced Spatial Cross-Attention block (SSCA), which are stacked together to form a SliceOcc layer. After passing through several SliceOcc layers, the slice queries are refined and connections are established with the image features and neighboring planes.


\textbf{Slice Queries.} SliceOcc utilizes multiple pairs of slice queries to enable interaction between points on planes and their surrounding spatial regions. As introduced in the previous subsection, we slice the scene into $S$ parts. For the $i$-th slice level, we define a floor slice query $Q^{\text{floor}}_{i}$ (simplified as $Q^{f}_{i}$) and a ceiling slice query $Q^{\text{ceiling}}_{i}$ ($Q^{c}_{i}$) responsible for querying the lower and upper portions of the partial scene, respectively. Each slice query consists of a content query and a height query. We project the 3D grid anchors $\mathcal{A}$ onto the image features output by the encoder and use the sampled features to initialize the content query. The height query is a learnable embedding designed to capture the geometric information at different slice levels.


\vspace{-10pt}
\begin{equation}
    \mathbf{Q} = \bigcup_{i}^{S} \{ ( \mathbf{Q}^{\text{floor}}_{i}, \mathbf{Q}^{\text{ceiling}}_{i} ) \}, \quad i \in [1, S]
\end{equation}
\vspace{-10pt}
\begin{equation}
    \mathbf{Q}^{j}_{i} = Q^{\text{content}, j}_{i} + Q^{\text{height}}_{i}, \quad j \in \{\text{floor}, \text{ceiling}\}
\end{equation}
\vspace{-10pt}
\begin{equation}
    Q^{\text{content}, j}_{i} = \mathcal{F}(\pi(\mathcal{A}_{i}))
\end{equation}

\noindent where $\pi$ denotes the anchor projection process, $\mathcal{A}_{i} \in \mathcal{A}$ represents the $i$-th slice level anchor points.

\textbf{Planar Cross Attention (PCA).} Due to the fact that objects in indoor scenes exhibit a certain consistency in height, we infer that neighbored planar slice features also tend to display strong local correlations. To leverage this geometric characteristic of the scene, we propose planar cross attention to interact with the correlated slice features between neighbored planes. We construct PCA with deformable attention \cite{zhu2020deformable}.

For a pair of slice queries $Q^c_i$ and $Q^f_i$ at the $i$-th slice level, we first obtain a query token $q^{f}_{i,x,y}$ at the $(x,y)$ location position from $Q^f_i$, and define reference points for  $q^{f}_{i,x,y}$ to calculate deformable attention of features on the other plane. Since 
$Q^c_i$ and $Q^f_i$ have the same spatial shape, we directly define two sets of grid points evenly distributed along the orthogonal horizontal axes and stack them together to form planar reference points  $\mathbf{Ref}^{2d} \in \mathbb{R}^{W\times L}$. Subsequently, the first step of the PCA block is implemented through cross attention between $q^{f}_{i,x,y}$ and $Q^c_i$:
\vspace{-5pt}
\begin{equation}
    \text{PCA}(q^f_{i,x,y},Q_i^c) =   \text{D.A.}(q^f_{i,x,y}, \mathbf{Ref}^{2d}, Q_i^c)
\end{equation}

\noindent where $D.A.$ represents multi-scale deformable attention \cite{zhu2020deformable}. After the first stage of PCA, $Q^f_i$ gathers knowledge from $Q^c_i$ and is updated through cross attention with $Q^c_i$. Then, we use tokens $q^c_{i,x,y}$ from $Q^c_i$ as queries and tokens from the updated $Q^f_i$ as values for the second stage of PCA:
\vspace{-4pt}
\begin{equation}
    \text{PCA}(q^c_{i,x,y},Q_i^f) = \text{D.A.}(q^c_{i,x,y}, \mathbf{Ref}^{2d}, Q_i^f)
\end{equation}

\textbf{Sliced Spatial Cross-Attention (SSCA).} To enable planar queries to interact with neighboring points in 3D space and further extract useful information from images, we propose the sliced spatial cross attention block (SSCA). We also adopt deformable attention to construct the SSCA due to the efficient implementation.

For a (ceiling or floor) slice query token $q_{i,x,y}$ at query location $(x, y)$ of the $i$-th slice level, we first define a set of 3D reference points to help establish a mapping relationship between the token and surrounding spatial points. For the ceiling query, we set pillar-shaped grid points extending downward from the ceiling, with a length equal to half of the scene's height. For the floor query, we define similar pillars extending upward from the floor at the same slice level. The 3d reference points $\mathbf{Ref}_i^{3d}$ for $q_{i,x,y}$ is described as:
\vspace{-3pt}
\begin{equation}
    (x', y') = ((x - \frac{L}{2}) p^h, (y - \frac{W}{2}) p^w)
\end{equation}
\vspace{-10pt}
\begin{equation}
    \mathbf{Ref}_i^{3d} = \{ (x', y', \mathbf{z_{i}}) \}
\end{equation}
\vspace{-10pt}
\begin{equation}
     \mathbf{z_{i}} = \{ (i-1+\beta )(\frac{H}{S}) + n(\frac{H}{2SN^{r3d}}) \}
\end{equation}


\noindent where $W$, $L$ denotes the spatial resolution of slice queries; $p^h$, $p^w$ is range of horizontal scene boundaries, $\beta \in \{0, \frac{1}{2}\}$, $N^{r3d}$ is the number of 3D reference points within a pillow, $H$ is the scene height as previously defined, and $n \in [1, N^{r3d}]$ represents the order of reference points. 

Subsequently, we project the 3D reference points $\mathbf{Ref}_i^{3d}$ through the camera intrinsics and extrinsics onto the $j$-th image plane to obtain reference points on the image features.
\vspace{-6pt}
\begin{equation}
    \mathbf{Ref}_{i,j}^{3d \xrightarrow{}2d} = \mathbf{K_j} \cdot \mathbf{E_j} \cdot \mathbf{Ref}_i^{3d} 
\end{equation}

Following this, based on the reference points projected onto the image $\mathbf{Ref}_i^{3d \xrightarrow{}2d}$, we can compute the deformable attention between the slice query and the image features $\mathcal{F}$.
\vspace{-3pt}
\begin{equation}
    \footnotesize
    \text{SSCA}(q_{i,x,y},\mathcal{F}) = \frac{1}{|\mathcal{V}_{hit}|}\sum_{j \in \mathcal{V}_{hit}} \text{D.A.}(q_{i,x,y}, \mathbf{Ref}_{i,j}^{3d \xrightarrow{}2d}, \mathcal{F}_j)
\end{equation}

\noindent where $\mathcal{V}_{hit}$ denotes the set of views where $\mathbf{Ref}_i^{3d \xrightarrow{}2d}$ are effectively projected onto the image plane.

\subsection{3D semantic occupancy prediction.}
After obtaining $S$ pairs of slice features, our goal is to estimate the 3D semantic occupancy of any 3D grid center in the space. When given a center point $p$ with its grid index $(x, y, z)$ in the space, we achieve this by obtaining the planar feature projection of $p$ through interpolating its neighboring ceiling slice feature $Q_s^c$ and floor slice feature $Q_s^f$ using tri-linear methods. We concatenate the $S$ pairs of grid features along the vertical direction to obtain the scene's voxel features. Finally, we use a 3D FCN (Fully Convolutional Network) head on voxel features to obtain the semantic probability for each occupancy grid. 
\subsection{Loss function.}
In terms of model training, we adopt the Scene-Class Affinity Loss from MonoScene \cite{cao2022monoscene}, dividing the supervision signals into geometric and semantic aspects. For the geometric aspect, we construct the loss $L_{geo} = L_{scal}(\hat{\mathbf{x}}^{geo}, \mathbf{y}^{geo})$ by comparing the model's predicted probability $\hat{\mathbf{x}}^{geo} \in \mathbb{R}^{1\times W \times L \times H}$ of grids being non-empty (occupied) with the binary labels for all semantic classes in the label $\mathbf{y}^{geo} \in \mathbb{R}^{1\times W \times L \times H}$. As for semantic supervision, given the model's predicted semantic probabilities $\hat{\mathbf{x}}^{sem} \in \mathbb{R}^{C \times W \times L \times H}$ for the grids and the corresponding semantic labels $\mathbf{y}^{sem} \in \mathbb{R}^{1\times W \times L \times H}$, we can construct the following semantic loss as $L_{sem} = L_{scal}(\hat{\mathbf{x}}^{sem}, \mathbf{y}^{sem})$. Besides, we also introduce a standard cross-entropy loss $L_{ce}$. Finally, SliceOcc is trained end-to-end with the following loss function:
\vspace{-5pt}
\begin{equation}
    L_{total} = L_{ce} + L_{geo} + L_{sem}
\end{equation}

\begin{table*}[ht]
    \centering
    \resizebox{0.95\linewidth}{!}{
    \begin{tabular}{c|c|c|cccccccccccc}
        \toprule
        \multirow{2}{*}{Method} & \multirow{2}{*}{Input} & \multirow{2}{*}{mIoU} 
        & \rotatebox{90}{empty} & \rotatebox{90}{floor} & \rotatebox{90}{wall} & \rotatebox{90}{chair} &  \rotatebox{90}{cabinet}& \rotatebox{90}{door} & \rotatebox{90}{table} & \rotatebox{90}{couch} & \rotatebox{90}{shelf} & \rotatebox{90}{window} & \rotatebox{90}{bed}  \\
        & & & \textcolor{white}{$\blacksquare$} & \textcolor{orange}{$\blacksquare$} & \textcolor{pink}{$\blacksquare$} & \textcolor{yellow}{$\blacksquare$} & \textcolor{blue}{$\blacksquare$} & \textcolor{cyan}{$\blacksquare$} & \textcolor{green}{$\blacksquare$} & \textcolor{red}{$\blacksquare$} & \textcolor{gray}{$\blacksquare$} & \textcolor{darkbrown}{$\blacksquare$} & \textcolor{purple}{$\blacksquare$}  \\   
        \midrule
            SPVCNN & Depth & 7.32 & 63.04 & 61.30 & 38.82 & 25.10 & 15.28 & 7.55 & 26.60 & 16.23 & 18.19 & 8.26 & 25.64 \\          
        Asym-UNet & Depth & 11.52 & \textbf{70.22} & \textbf{63.53} & 44.18 & 41.54 & 20.63 & 17.35 & 34.02 & 35.67 & 26.48 & 15.98 & \textbf{42.89}  \\         
        Mink-ResNet34 & Depth & \textbf{15.56} & 69.92 & 60.52& \textbf{51.74} & \textbf{49.44} & \textbf{23.08} & \textbf{24.33} & \textbf{45.77} & \textbf{43.52} & \textbf{29.74} & \textbf{23.02} &39.04\\
        \midrule
        OccNet & RGB & 8.07 & 37.15 & 46.90 & 25.63 & 20.94 & 13.17 & 18.40 & 26.81 & 22.86 & 13.59 & 13.49 & 26.75 \\
        SurroundOcc & RGB & 9.10 & 38.54 & 46.17 & 23.55 & 23.04 & 13.60 & 19.15 & 27.79 & 22.28 & 13.11 & 13.72 & 24.32  \\
        EmbodiedScan & RGB & 10.48 &40.45 &41.25 &27.19 &26.16 &15.50 &20.30 &30.82 &26.70 &15.01 &14.33 &29.17  \\
        $\text{EmbodiedScan}^{\dagger}$ & RGB & 14.52 & 48.81 & 49.57 & 32.32 & \textbf{32.12} & 20.95 & 24.30 & 27.95 & 30.51 & \textbf{27.30} & 19.03 & 35.40  \\
 
        SliceOcc (ours) & RGB & \textbf{15.45} & \textbf{49.32} & \textbf{51.39} & \textbf{33.29} & 31.65 & \textbf{22.39} & \textbf{25.14} & \textbf{29.57} & \textbf{32.78} & 26.83 & \textbf{19.88} & \textbf{38.88}  \\
        \bottomrule
    \end{tabular}
    }
    \caption{Quantitative results on EmbodiedScan multi-view occupancy prediction benchmark. We selected \textbf{11 common categories} out of the total 81 categories in the dataset for analysis. For a fair comparison, we re-implement EmbodiedScan as $\text{EmbodiedScan}^{\dagger}$, \text{optimized with the same input view number as ours.} The best results within each modality are bolded.}
    \label{tab:exp}
\end{table*}
\vspace{-10pt}
\begin{figure*}[htbp]
\centering
\begin{subfigure}[b]{0.15\textwidth}
\includegraphics[width=\textwidth]{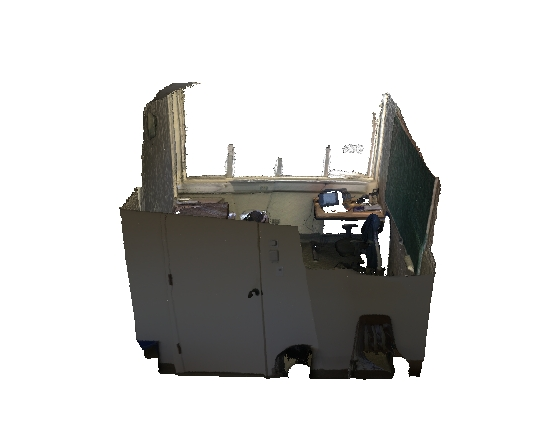}
\label{fig:subfig1}
\end{subfigure}%
\hspace{0.05\textwidth}
\begin{subfigure}[b]{0.15\textwidth}
\includegraphics[width=\textwidth]{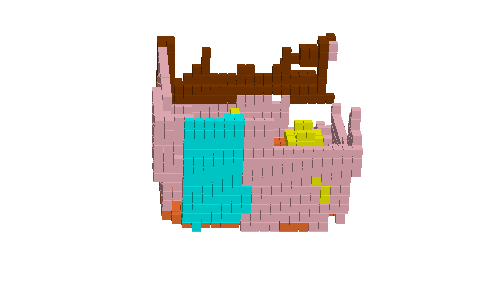}
\label{fig:subfig2}
\end{subfigure}%
\hspace{0.05\textwidth}
\begin{subfigure}[b]{0.15\textwidth}
\includegraphics[width=\textwidth]{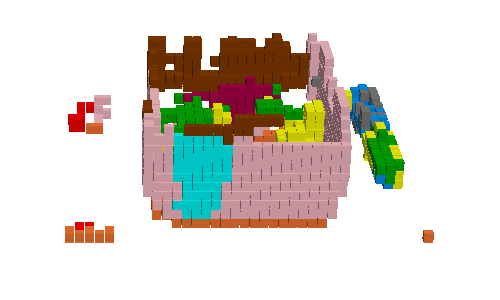}
\label{fig:subfig3}
\end{subfigure}%
\hspace{0.05\textwidth}
\begin{subfigure}[b]{0.15\textwidth}
\includegraphics[width=\textwidth]{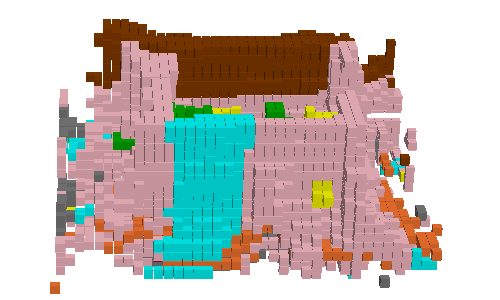}
\label{fig:subfig4}
\end{subfigure}
\hspace{0.05\textwidth}
\begin{subfigure}[b]{0.15\textwidth}
\includegraphics[width=\textwidth]{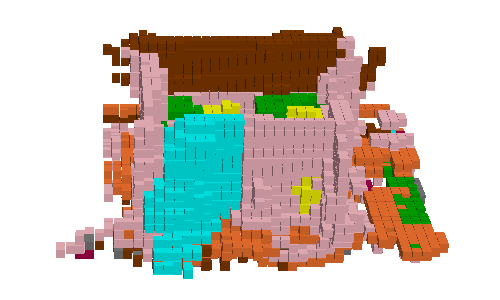}
\label{fig:subfig5}
\end{subfigure}

\vspace{-1.3\baselineskip}
\begin{subfigure}[b]{0.15\textwidth}
\includegraphics[width=\textwidth]{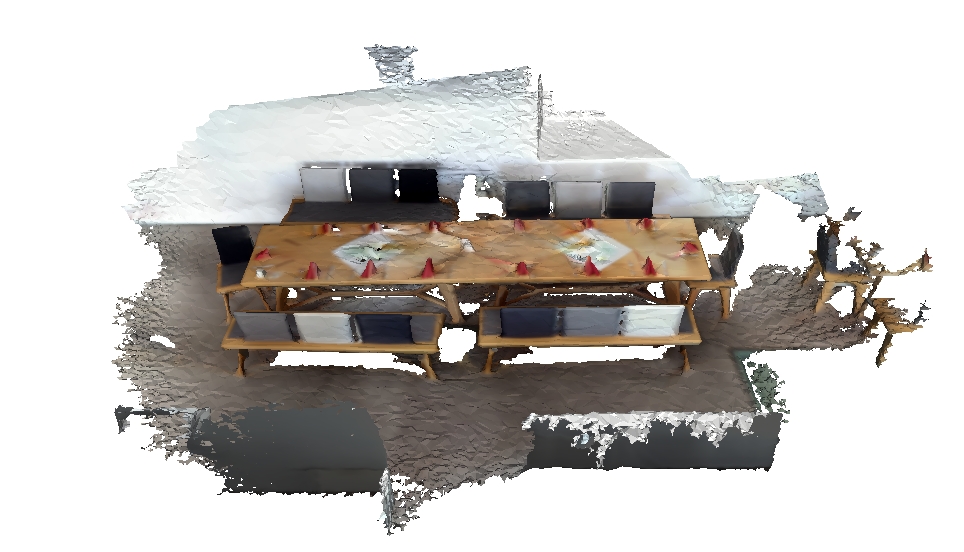}
\label{fig:subfig1}
\end{subfigure}%
\hspace{0.05\textwidth}
\begin{subfigure}[b]{0.15\textwidth}
\includegraphics[width=\textwidth]{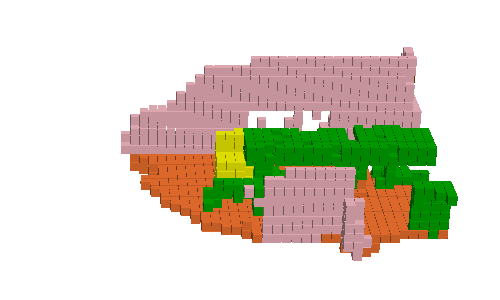}
\label{fig:subfig2}
\end{subfigure}%
\hspace{0.05\textwidth}
\begin{subfigure}[b]{0.15\textwidth}
\includegraphics[width=\textwidth]{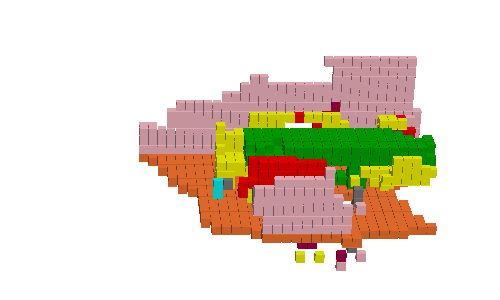}
\label{fig:subfig3}
\end{subfigure}%
\hspace{0.05\textwidth}
\begin{subfigure}[b]{0.15\textwidth}
\includegraphics[width=\textwidth]{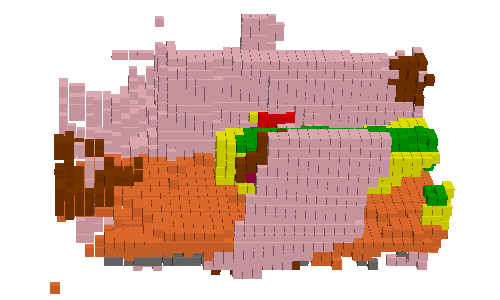}
\label{fig:subfig4}
\end{subfigure}
\hspace{0.05\textwidth}
\begin{subfigure}[b]{0.15\textwidth}
\includegraphics[width=\textwidth]{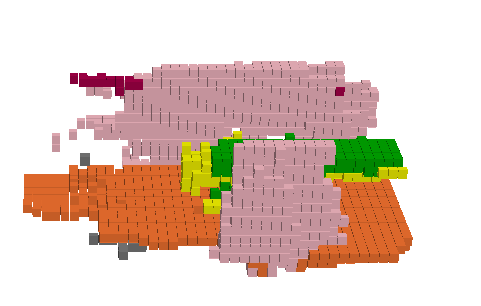}
\label{fig:subfig5}
\end{subfigure}

\vspace{-1.2\baselineskip}
\begin{subfigure}[b]{0.15\textwidth}
\includegraphics[width=\textwidth]{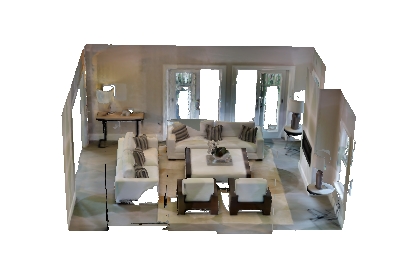}
\caption{Indoor Scene}
\label{fig:subfig1}
\end{subfigure}%
\hspace{0.05\textwidth}
\begin{subfigure}[b]{0.15\textwidth}
\includegraphics[width=\textwidth]{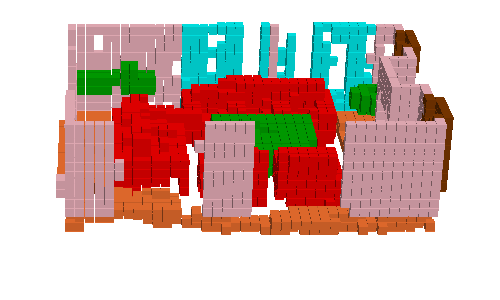}
\caption{GT}
\label{fig:subfig2}
\end{subfigure}%
\hspace{0.05\textwidth}
\begin{subfigure}[b]{0.15\textwidth}
\includegraphics[width=\textwidth]{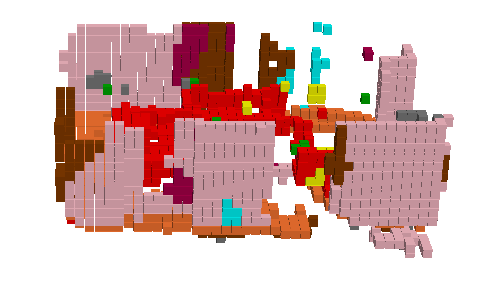}
\caption{Asym-UNet}
\label{fig:subfig3}
\end{subfigure}%
\hspace{0.05\textwidth}
\begin{subfigure}[b]{0.15\textwidth}
\includegraphics[width=\textwidth]{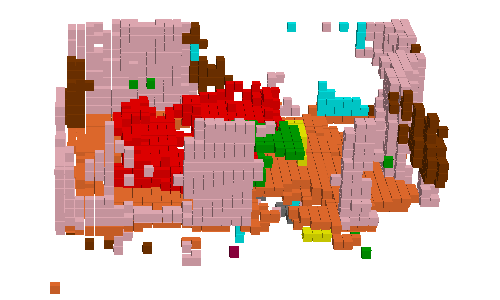}
\caption{EmbodiedScan}
\label{fig:subfig4}
\end{subfigure}
\hspace{0.05\textwidth}
\begin{subfigure}[b]{0.15\textwidth}
\includegraphics[width=\textwidth]{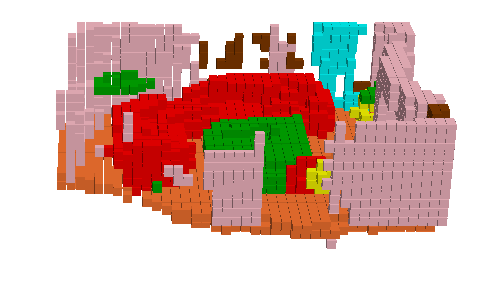}
\caption{SliceOcc}
\label{fig:subfig5}
\end{subfigure}
\caption{Qualitative performance on the EmbodiedScan multi-view occupancy prediction benchmark.}
\label{fig:visualization}
\end{figure*}

\section{Experiments}
\subsection{Datasets}
Following EmbodiedScan \cite{wang2024embodiedscan}, we selected three indoor scene datasets, specifically ScanNet v2 \cite{dai2017scannet}, Matterport3D \cite{chang2017matterport3d}, and 3RScan \cite{wald2019rio}, to conduct experiments on 3D semantic occupancy prediction. \textbf{ScanNetv2} is a large-scale RGB-D dataset that supports indoor 3D scene understanding, containing 1,513 manually annotated indoor scenes with over 2.5M views. \textbf{Matterport3D} contains 90 indoor building scenes, featuring over 194K RGB-D views. \textbf{3RScan} covers 478 indoor scenes, providing 1,482 scene scans.
Ultimately, we had a total of 3,112 training scenes and 389 testing scenes and utilized the semantic occupancy labels from EmbodiedScan for supervised training.
\subsection{Implementation details}

In terms of model architecture, we employed a ResNet-50 as the image encoder. The sampling number of input images $N$ was set to 20. In the occupancy decoder, SliceOcc consisted of three transformer layers, each of which includes a PCA block followed by a SSCA block. The slice resolution $W\times L$ was fixed at $40\times 40$. The slice number $S$ was set to 16, and the 3d reference point number $N^{r3d}$ was set to 4 as default. In data processing, the input images were resized to $480\times480$. The scene boundaries were set to [-3.2m, 3.2m] for both length and width, and [-1.28m, 1.28m] for height. The shape of the occupancy labels and the final output of the model were both $40\times40\times16$. For training, we utilized the AdamW optimizer with a learning rate of 1e-4 and a weight decay rate of 1e-2.  All experiments were conducted on 8 RTX 4090 GPUs.

\subsection{Baselines}
We selected two types of models as baselines: RGB camera-based and depth-based models. RGB camera-based models used multi-view images as input and establish a mapping relationship from 2D images to 3D space to estimate the scene's semantic occupancy.  By contrast, depth-based models used point clouds as input, which inherently possessed dense 3D geometric properties. For \textbf{RGB} camera-based models, we selected OccNet \cite{tong2023scene}, SurroundOcc \cite{wei2023surroundocc}, and EmbodiedScan \cite{wang2024embodiedscan} as baselines. For \textbf{depth}-based models, we chose SPVCNN from SPVNAS \cite{tang2020searching}, Asym-UNet from Cylinder3D \cite{zhu2021cylindrical}, and Mink-ResNet34 from MinkowskiNet \cite{choy20194d} as baseline models. When re-implementing the depth-based methods, we used the same experimental settings as the depth-only model in EmbodiedScan.

\begin{table*}[htbp]
\centering
\begin{minipage}{0.30\textwidth}
\centering
\scalebox{0.9}{
\begin{tabular}{c|c|c@{\,}c}
\toprule
\multirow{2}{*}{\makecell{\textbf{Layer} \\ \textbf{Number}}} & \textbf{mIoU} & \textbf{FLOPs} & \#\textbf{Par.}\\
                                                     & (\%)          & (G)      & (M)       \\    
\midrule
1                                                      & 14.63         & 902         & 120     \\
2                                                      & 15.33         & 989        & 128       \\
3                                                      & 15.45         & 1107        & 137       \\
\bottomrule
\end{tabular}
}
\caption{Analysis of variations in the number of SliceOcc layers.}
\label{tab:ab1}
\end{minipage}
\hfill
\begin{minipage}{0.30\textwidth}
\centering
\scalebox{0.9}{
\begin{tabular}{c|c|c@{\,}c}
\toprule
\multirow{2}{*}{\makecell{\textbf{Slice} \\ \textbf{Number}}} & \textbf{mIoU} & \textbf{FLOPs} & \#\textbf{Par.}\\
                                                     & (\%)          & (G)      & (M)       \\    
\midrule
2                                                      & 12.13         & 903      & 110         \\
4                                                      & 13.40         & 926      & 114       \\
8                                                      & 14.57         & 970     & 121               \\
\bottomrule
\end{tabular}
}
\caption{Effect of different slice numbers on model performance.}
\label{tab:ab2}
\end{minipage}
\hfill
\begin{minipage}{0.325\textwidth}
\centering
\scalebox{0.9}{
\begin{tabular}{c|c|c@{\,}c}
\toprule
\multirow{2}{*}{\makecell{\textbf{Slice} \\ \textbf{Resolution}}} & \textbf{mIoU} & \textbf{FLOPs} & \#\textbf{Par.} \\
                                                     & (\%)          & (G)      & (M)       \\    
\midrule
$20\times20$                                          & 14.06         & 927    & 132           \\
$30\times30$                                          & 13.64         & 960    & 134           \\
$35\times35$                                          & 14.68         & 1022    & 136          \\
\bottomrule
\end{tabular}
}
\caption{Comparison of performance at different slice resolutions.}
\label{tab:ab3}
\end{minipage}

\end{table*}

\begin{figure*}[h]
\centering
\begin{minipage}{0.32\textwidth}
\centering
\includegraphics[width=0.8\textwidth]{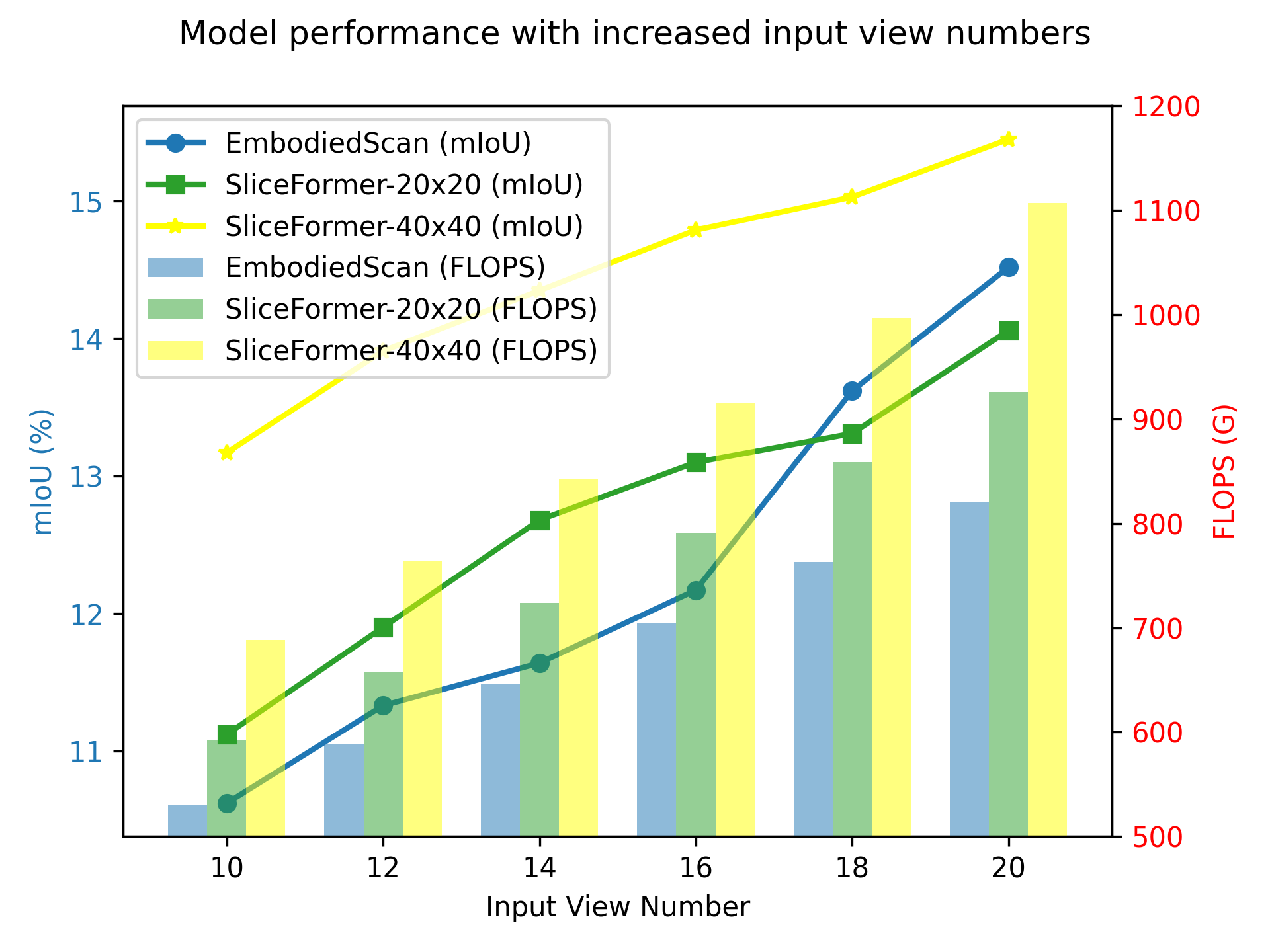}
\caption{Effect of increasing input views on SliceOcc performance.}
\label{fig:ab1}
\end{minipage}
\hfill
\begin{minipage}{0.32\textwidth}
\centering
\includegraphics[width=0.8\textwidth]{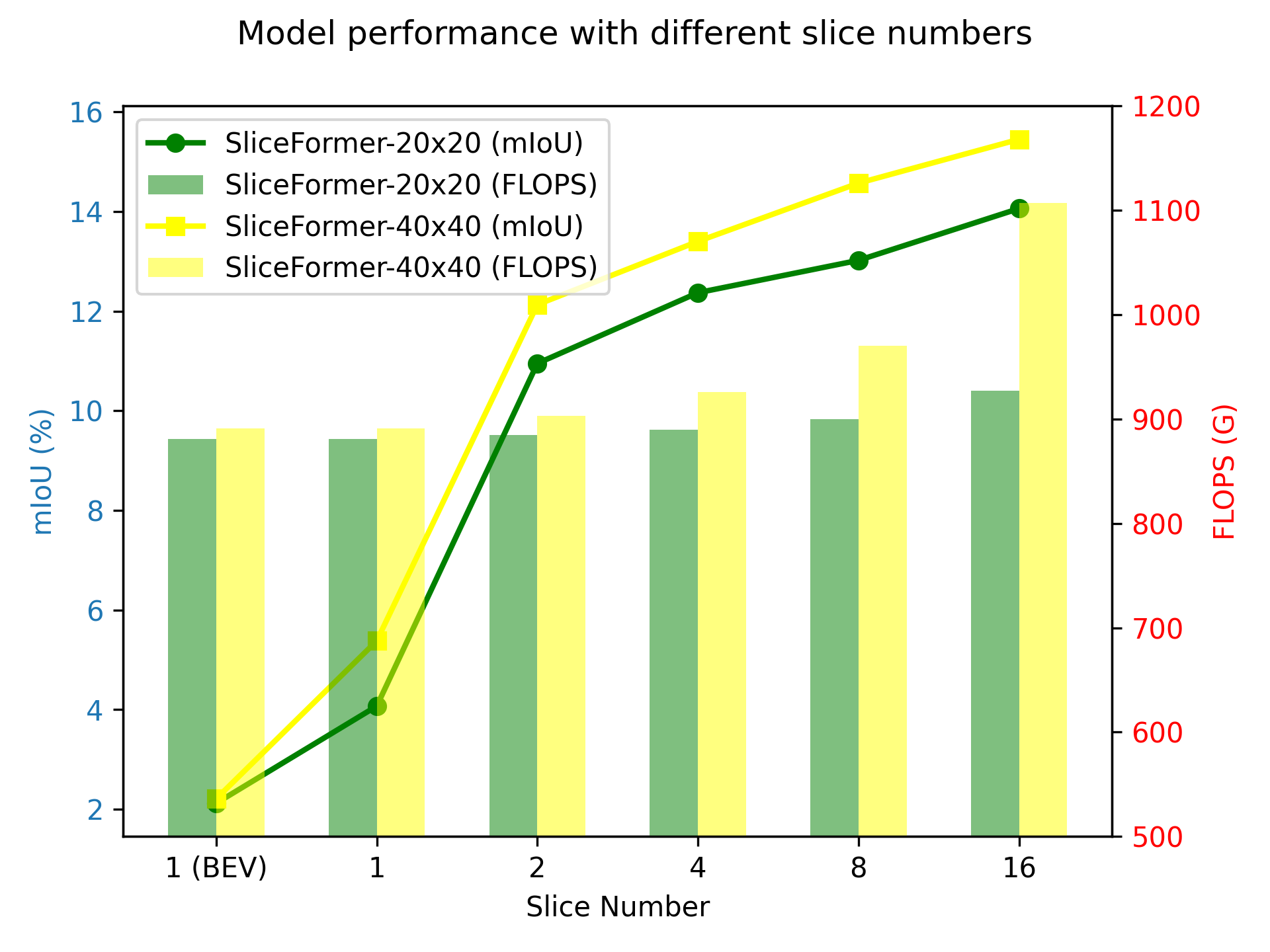}
\caption{Additional statistics on the effect of varying slice number.}
\label{fig:ab2}
\end{minipage}
\hfill
\begin{minipage}{0.32\textwidth}
\centering
\includegraphics[width=0.85\textwidth]{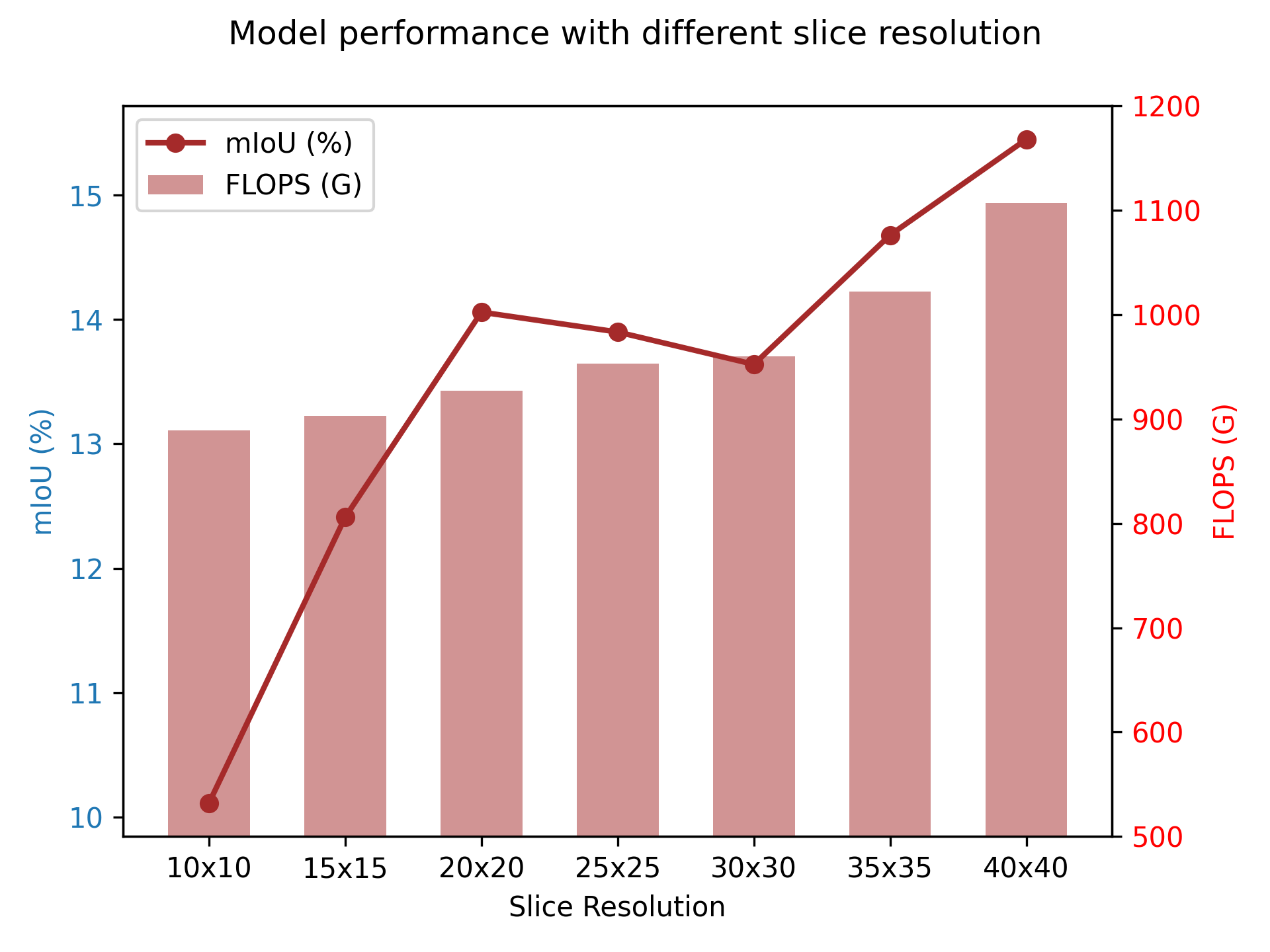}
\caption{Additional performance record on the effect of different slice resolution.}
\label{fig:ab3}
\end{minipage}
\end{figure*}

\subsection{Performance}
As shown in Table \ref{tab:exp}, among RGB camera-based models, SliceOcc significantly outperformed autonomous driving occupancy models such as OCCNet and SurroundOcc in mIoU. Compared to the state-of-the-art indoor occupancy model, EmbodiedScan, we also demonstrated a notable advantage. Additionally, we achieved performance comparable to popular depth-based models. In terms of mIoU, our performance was very close to Mink-ResNet34 and exceeded it in some categories. Overall, we achieved the best results in 39 out of 81 categories compared to the listed baselines.

In Fig. \ref{fig:visualization}, we provided visualization results. For all tested scenes, we selected one scene each from the ScanNet (top row), 3RScan (middle row), and Matterport3D (bottom row) datasets for qualitative analysis. It can be seen that depth-based methods (Asym-UNet) accurately modeled global scene geometry but were slightly inferior to RGB camera-based methods (EmbodiedScan and SliceOcc) in local semantic prediction. Our SliceOcc was capable of clearly predicting the scene's geometric structure and the semantics of local objects, while also demonstrating a higher signal-to-noise ratio (SNR) than EmbodiedScan. The color mapping of the qualitative results can be found in Table \ref{tab:exp}.

\begin{table}[t]
    \centering
    \scalebox{0.9}{
    \begin{tabular}{c|ccc|c}
    \toprule
    \multirow{2}{*}{\textbf{Type}} & \multirow{2}{*}{\textbf{SSCA}} & \multirow{2}{*}{\textbf{PCA}} & \multirow{2}{*}{\makecell{\textbf{Height} \\ \textbf{query}}}& \textbf{mIoU}  \\
                                                         &     &   &  & (\%)          \\    
    \midrule
     \textit{base}   & \textcolor{red}{-} &  \textcolor{red}{-}  &     \textcolor{red}{-}    & 14.13        \\
       I                    & \textcolor{green}{\checkmark}  & \textcolor{red}{-}   &  \textcolor{red}{-}     & 15.12         \\
       II                   &  \textcolor{red}{-} &  \textcolor{green}{\checkmark}  &     \textcolor{red}{-}   & 14.70          \\
       III                    &  \textcolor{green}{\checkmark} &  \textcolor{green}{\checkmark}  &   \textcolor{red}{-}   & 15.36        \\
       IV                    &  \textcolor{green}{\checkmark} &   \textcolor{green}{\checkmark} &  \textcolor{green}{\checkmark}     & 15.45          \\
    \bottomrule
    \end{tabular}}
    \caption{Ablating the effect of different components.}
    \label{tab:sup1}
\end{table}

\subsection{Experimental Analysis and Ablation Study}

\noindent \textbf{Input views.} The number of input views is also a crucial factor affecting the model's performance, as changes in the number of views can impact the model's effective perception of occupancy through factors such as field of view and occlusion. We kept the number of input views consistent between training and testing process and increased the number of views from 10 to 20. We comprehensively evaluated EmbodiedScan, as well as SliceOcc with slice resolutions of $20\times20$ and $40\times40$, obtaining the results in Fig. \ref{fig:ab1}.

\noindent \textbf{Layer number.} We conducted analysis on the number of transformer layers, where each transformer layer includes a PCA block, a SSCA block, and a FFN. As shown in Table \ref{tab:ab1}, The model's performance, along with its computational cost (FLOPs) and number of parameter (\#Par.) increases with the number of layers. Specifically, increasing layer number from 1 to 2  brings a 0.7\% improvement in mIoU. 

\begin{figure}[h]
    \centering
    \includegraphics[width=0.8\linewidth]{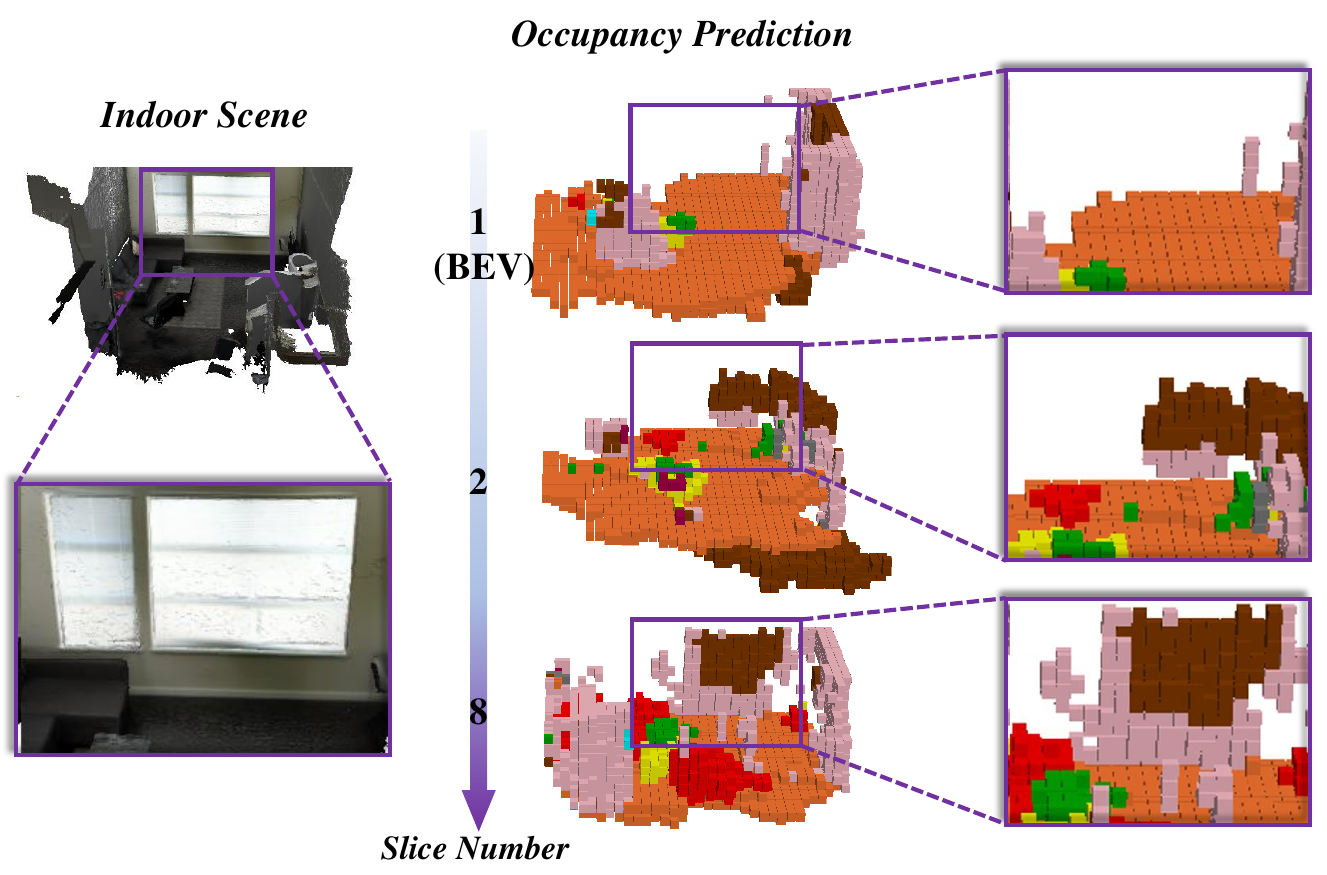}
    \caption{As the slice number increases, SliceOcc outputs more accurate predictions in both geometric and semantic aspects.}
    \label{fig:increaseslice}
\end{figure}

\noindent \textbf{Slice numbers.} We changed the slice number used to divide the scene and track the model performance. The slice number affects the model's ability to perceive fine-grained details in the vertical direction, as well as its capability to handle the stacking (occlusion) of objects. In Table \ref{tab:ab2}, the model performance is reported for slice numbers of 2, 4, and 8. To further highlight the advantages of the slice representation, we also conducted experiments with a single plane (BEV), where only the floor slice is used to reconstruct the voxel features when the slice number is 1. Additional experimental records were provided in Fig. \ref{fig:ab2}. We also provided visualizations demonstrating how the model's predictions improve with an increasing number of slices in Fig. \ref{fig:increaseslice}.

\noindent \textbf{Slice resolution.} We compared the performance of SliceOcc with different slice resolutions. As recorded in Table \ref{tab:ab3}, we chose $20\times20$, $30\times30$, and $35\times35$ as different slice resolutions, which correspond to the spatial shapes of the slice queries. We found that the model's performance does not consistently increase with the growth of slice resolution. For example, increasing the slice resolution from $20\times20$ to $30\times30$ results in a change in mIoU of -0.42\%. To more comprehensively illustrate this trend, we present additional analysis regarding slice resolution in Fig. \ref{fig:ab3}.


\noindent \textbf{Ablation of different components.} We evaluated the effect of key components in SliceOcc. Introducing the SSCA and PCA modules separately leads to Type I (15.12\%) and Type II (14.70\%) results, as shown in Table \ref{tab:sup1}
. When both SSCA and PCA are combined, the prediction performance improves to 15.36\%. Additionally, incorporating a height query into the slice query further enhances accuracy, reaching 15.45\%.


\vspace{-10pt}
\section{Conclusion}
This paper introduces SliceOcc, an innovative planar-based model tailored for 3D semantic occupancy prediction in indoor environments. By utilizing a vertical slice representation and leveraging cross-attention mechanisms, SliceOcc effectively captures the intricate details of indoor scenes, particularly addressing the challenges posed by dense occlusions. This capability could potentially benefit robotic vision systems, allowing for more accurate perception and navigation in cluttered indoor spaces. Extensive experiments conducted on the EmbodiedScan dataset validate the efficacy of our approach, with SliceOcc achieving a state-of-the-art mIoU of 15.45\% across 81 object categories. These results highlight the model's ability to outperform existing RGB camera-based approaches and match the performance of leading depth-based methods, suggesting its potential as a useful tool for advancing indoor 3D scene understanding.

{
\bibliographystyle{IEEEtran}
\bibliography{IEEEabrv,reference}
}
\end{document}